\documentclass[12pt,a4paper]{article}

\usepackage{amsmath}
\usepackage{algorithm}
\usepackage[noend]{algpseudocode}
\usepackage{url}
\usepackage{hyperref}
\usepackage{graphicx}

\begin{document}
\title{Gradient boosting with vector-valued leafs}
\author{David Cortes}
\maketitle

\begin{abstract}
Gradient boosting in the form of decision tree ensembles has successfully been applied to a variety of problems using simple objective functions based on log-likelihoods of a single variable. The concept extends naturally to objective functions operating on vectors - for example, multinomial logistic log-likelihood for multi-class classification, where observations have a score for each class - but popular frameworks approach these functions by either updating one value of the input vectors at a time, or by using a diagonal upper bound on the second derivative. This work extends the usual gradient boosting framework to functions of vector inputs and sketches a simple algorithm that can be used efficiently with histogram-based decision trees.
\end{abstract}

\section{Introduction}
One of the most popular frameworks for statistical modeling is gradient boosting in the form of additive decision tree ensembles (also known as GBT for 'gradient boosted trees'), with some of the most popular software implementations being XGBoost (\cite{xgb}), LightGBM (\cite{lgb}), and CatBoost (\cite{catboost}). All three of these libraries offer a variety of so-called objective functions to minimize based on different types of decision-tree models fitted to observed data, with many of these functions being derived from log-likelihoods commonly used in GLMs (generalized linear models), including logistic log-likelihood, gamma, Poisson, among many others.

Initially, the theoretical framework introduced in \cite{xgb} considered objectives that consist of additive functions (over i.i.d. observations) of one variable each, but these have later been extended to other (additive) objective functions that operate on vectors as inputs, such as multinomial logistic log-likelihood for multi-class classification, where each observation would need to get a different score for each class based on rules dervived from the same features.

The initial introduction of these objectives was done using the same framework, by updating the scores for one component of the vector of inputs at a time - e.g. updating the predictions for one class while holding the predictions for other classes constant, by making each leaf in the tree contain scores for a single class. Later on, some of these libraries have begun offering a different logic in which vector-valued objectives are approached through vector updates - e.g. each leaf in a tree has scores for each class - but these are based on an approximation of the second derivative (Hessian) for both the split gain calculation and the leaf values, which is less optimal than using the true Hessian of the objectives to minimize.

This work provides an overview of gradient boosting with vector-valued functions using the true Hessian, and proposes an algorithm that can be used efficiently in histogram-based decision tree routines.

\section{Overview of gradient boosting}
In gradient boosting, the idea is to minimize a convex function by making additive updates based on sequential quadratic approximations, given by the first and second derivates of the function, also known as the gradient and Hessian.

Mathematically, given a set of $n$ observations with a known and fixed response value $y$, the goal is to assign scores $x$ to different observations, in such a way that they minimize an additive objective function:
$$
\text{argmin}_{x_{1..n}} \sum_{i=1}^n f(x_i, y_i)
$$

In a decision tree, the scores $x_{i..n}$ are constrained to be the same constant for each observation meeting criteria of the form "if feature1 is less than 't' and feature2 is more than 'z', then score is 'w'", with these rules being derived from binary decision trees constructed from top to bottom which group observations together into branches and leafs according to conditions based on the values of the input features / covariates.

In the simplest case, if the function to minimize is squared error:
$$
f(x_i, y_i) = (y_i - x_i)^2
$$

The optimal score to assign as $x_i$ to all the observations that fall into the same bucket / leaf can be calculated in closed form, as the mean of the response:
$$
x^{*}_{\text{leaf}} = \frac{1}{|S|} \sum_{i \in S} y_i = \bar{y}
$$

Where $S$ is the subset of observations belonging to the same bucket, and $|S|$ denotes its cardinality. This makes squared error a useful base for the rest of the framework.

Although it is generally not known apriori what the optimal assignment of observations to buckets / leafs would be, for squared loss, if a decision tree is constrained to consist of a single decision rule splitting observations among two branches, then the optimal split will be the one that minimizes the sum of squared residuals among both branches if both of them get assigned the optimal score given the observations they contain:
$$
\min \sum_{i \in L} (y_i - x^{*}_l)^2 + \sum_{i \in R} (y_i - x^{*}_r)^2
$$

Where $L$ and $R$ denote the subsets of observations belonging to the left brach ("feature is less than 'y'") and right branch ("feature is more than 'y'"), respectively. Under squared error, minimizing this quantity is equivalent to maximizing the information gain criterion:
$$
\max \frac{
    \text{Var}(y_{\text{all}}) - \frac{|L|}{n}\text{Var}(y_l) - \frac{|R|}{n}\text{Var}(y_r)
}{ \text{Var}(y_{\text{all}}) }
$$

Where $\text{Var}(y)$ denotes the variance of the response variable, among different subsets.

Note that:
$$
n \text{Var}(y) = n (\frac{1}{n} \sum_{i=1}^n (y_i - \bar{y}) )^2 = \text{rss}
$$

And since $n$ and $\text{Var}(y)$ are positive constants that do not depend on the data, maximizing the gain amounts to maximizing:
$$
\max - (\frac{|L|}{n}\text{Var}(y_l) + \frac{|R|}{n}\text{Var}(y_r) ) = - ( \sum_{i \in L} (y_i - \bar{y}_l)^2 + \sum_{i \in R} (y_i - \bar{y}_r)^2 )
$$

Which is equivalent to minimizing its negative.

For this simpler function, the optimal split point from which to make the decision rule can be calculated by looping over the data in sorted order of the feature from which the condition is made and evalauting the value of these quantities (minus constant terms) while keeping track of which split point provided the best value, calculating and updating only aggregates of the $y$ variable for each observation that would fall under each branch under each possible split point (given by the sorted order of the feature), and the optimal $x^{*}$ to assign to each branch / leaf can likewise be calculated from the same aggregates in the same pass. Note that the only required aggregates are sums - for the $y$ variable in this case.

As the optimal bucket / leaf assignments will not be known in advance, decision trees are built in a greedy manner making splits one at a time, from top to bottom, recursively. Note that the optimal split points and leaf values do not depend on the observations that are not part of the subset being evaluated, hence after one split, the same procedure can be repeated recursively and independently on the subsets that fell into each branch.

This procedure is typically accelerated by pre-grouping subsets of observations into buckets according to features (note that each observation has an independent additive effect), and evaluating the gain / squared residuals only on the split points between pre-grouped buckets. This is known as the histogram method. It then makes updates of aggregates faster (as long as the buckets do not change) as the aggregated quantities in one branch of a node can be calculated as the difference between the aggregations in the histogram of the parent branch vs. the other branch.

In the case of other objectives, since calculation of $x^{*}$ might not have a closed form and the optimal split point might not have an easy loopable form, gradient boosting operates on a second-order approximation of the function:
$$
f(x_{i..n} | x^0_{i..n}, y_{i..n}) \approx \sum_{i=1}^n (-g_i x_i + \frac{1}{2} h_i x_i^2) + k
$$

Where $g_i$ is the gradient (first partial derivative) of $f(x, y)$ evaluated at current $x^0_i$ (assume all $x_i$ are initialized to a constant), $h_i$ is the Hessian (second partial derivative) evaluated at current $x^0$, $x_i$ is the leaf values / scores that will be obtained from the decision tree and added to $x^0_i$, and $k$ is a constant that does not depend on $x_i$.

Under this formulation, if scores $x_i$ keep being summed from $t$ successive minimizations of this second-order approximation (such that $x^{\text{boost}}_i = \sum_{i=0}^t x^t_i$), the value of the original function $f(x^{\text{boost}}_i, y_i)$ is expected to decrease with successive iterations.

As such, GBT fits multiple additive, sequential decision trees based on successive approximations of these second-order approximations.

Note that the second order approximation corresponds to just a weighted version of squared error, with the weight for each observation given by $w_i = h_i$ and $\tilde{y}_i = \frac{g_i}{h_i}$. Its solution in a decision tree can be obtained with the same procedure described earlier, just with the difference that it involves sample weights that also need to be tracked in aggregates.

\section{GBT as matrix algebra}
The second-order minimization problem from gradient boosting for determining the optimal $x^{*}$ in a subset of $n$ observations can be expressed in matrix form as follows:
$$
\text{argmin}_{x} 0.5 \: \mathbf{x}^T \: \mathbf{H} \: \mathbf{x} - \mathbf{x}^T \mathbf{g}
$$

Where $\mathbf{H} = \text{diag}( \mathbf{h} )$ denotes a diagonal matrix of dimensions $n \times n$ having the values of vector $\mathbf{h}$ (the Hessians) in its main diagonal and zeros everywhere else, $\mathbf{x}$ is the (column) vector of scores / predictions, which in a decision tree leaf would need to be a scalar repeated $n$ times as a vector, and $\mathbf{g}$ is the vector of gradients.

Note again that this is a function of a scalar $x^{*}$, since $\mathbf{x}$ is constrained contain the same value for all observations falling into the same decision tree leaf.

Just like in IRLS (iteratively reweighted least squares), the function to minimize can be equivalently expressed as:
$$
\text{argmin}_{x} 0.5 \: \mathbf{x}^T \: \mathbf{H} \: \mathbf{x} -  \mathbf{x}^T \: \mathbf{H} \: ( \mathbf{H}^{-1} \mathbf{g} )
$$

One might see how this resembles the form of a weighted linear regression with covariates $\mathbf{X}$, weights $\mathbf{w}$ diagonalized as $\mathbf{W}$, and response $\mathbf{y}$:
$$
\beta^{*} = ( \mathbf{X}^T \mathbf{W} \mathbf{X} )^{-1} \mathbf{X}^T \mathbf{W} \mathbf{y}
$$

From this formula, it can be seen that the solution to $x^{*}$ is given by the solution to a linear regression of one variable (just an intercept), with $\mathbf{X} = \mathbf{1}^n$ (vector of ones), $\mathbf{W} = \mathbf{H}$, and $\mathbf{y} = \mathbf{H}^{-1} \mathbf{g}$.

The sum of squared residuals from this weighted linear regression likewise correspond to twice the function values from the second-order approximation, and recall again that maximizing the information gain is equivalent to minimizing the squared residuals:
$$
\text{rss} = || \mathbf{W}^{\frac{1}{2}} (\mathbf{y} - \mathbf{X} \beta) ||^2 = \sum_{i=1}^n ( h_i^\frac{1}{2} (\frac{g_i}{h_i} - \beta) )^2 = \sum_{i=1}^n 2 (0.5 h_i \beta^2 - g_i \beta) + \frac{g_i^2}{h_i}
$$

(where $\beta$ is the estimated $x^{*}$)

And note that the term $\frac{g_i^2}{h_i}$ does not depend on the $\beta$ to estimate, so it can be left out of the equation.

\section{Vector-valued functions}
Many functions of interest take on vector-valued inputs as $\mathbf{x}_i$ and/or $\mathbf{y}_i$. For example, multinomial logistic log-likelihood for multi-class classification for $k$ classes is a function of a vector $\text{x}_i$:
$$
f(\mathbf{x}, y) = -\log P(y = c) = -\log \frac{ \exp x_{c} }{ \sum_{i=1}^k \exp x_i }
$$

Where $c$ is the index of the class to which the observation belongs ($y_i$). In this case, the gradient $\mathbf{g}_i$ is also a vector of dimension $k$, and the Hessian $\mathbf{H}_i$ is a matrix of dimensions $k \times k$.

In \cite{vgam}, the GLM framework for scalar functions is extended to work with vector functions through the same IRLS reformulation trick, only with matrices of extended dimensions. The same formulae extend naturally to gradient boosting.

If extended to multple dimensions, the quadratic approximation around an initial vector $\mathbf{x}_{i..n}^0$ for each observation would become:
$$
f(\mathbf{x}_{i..n}) = \sum_{i=1}^n \frac{1}{2} \mathbf{x}_i^T \: \mathbf{H}_i \: \mathbf{x}_i - \mathbf{g}_i \: \mathbf{x}_i + \text{const}
$$

And recall again that a decision tree split should minimize the sum of this function across observations, with the constraint that all observations in the left branch would get the same $\mathbf{x}_{\text{left}}$, and all observations in the right branch would get the same $\mathbf{x}_{\text{right}}$.

Expressing this formula when all observations get the same $\mathbf{x}$ (what needs to be calculated for leaf values) through concatenated vectors, the objective to minimize in gradient boosting would be:
$$
\text{argmin}_{\mathbf{x}} 0.5 \: \mathbf{x}_{\text{long}} \: \mathbf{B} \: \mathbf{x}_{\text{long}} \: - \: \mathbf{x}_{\text{long}} \mathbf{g}
$$

But now $\mathbf{g}$ is a vector of dimension $k n$ made up by stacking the individual vector gradients (or equivalently, vectorizing a $k \times n$ matrix):
$$
\mathbf{g} = \begin{bmatrix} \:\mathbf{g}^k_1 \: \mathbf{g}^k_2 \:\: .. \:\: \mathbf{g}^k_n \: \end{bmatrix}
$$

$\mathbf{B}$ is a symmetric block matrix comprised of diagonal matrices of size $(n \times n)$ each:
$$
\mathbf{B} = \begin{bmatrix}

\mathbf{D}_{1,1} \: \mathbf{D}_{1,2} \:\: .. \:\: \mathbf{D}_{1,k} \\

\mathbf{D}_{2,1} \: \mathbf{D}_{2,2} \:\: .. \:\: \mathbf{D}_{2,k} \\

.. \\

\mathbf{D}_{k, 1} \: \mathbf{D}_{k, 2} \:\: .. \:\: \mathbf{D}_{k,k}

\end{bmatrix}
$$

With $\mathbf{D}_{p,q} = \mathbf{D}_{p,q}$ for non-diagonal elements, and each $\mathbf{D}_{p,q}$ consisting of a diagonal matrix made up of the elements $(p,q)$ of the Hessians $\mathbf{H}_i$ for each observation on the main diagonal and zeros elsewhere.

$\mathbf{x}_{\text{long}}$ is also made up by repeating the vector of length $k$ by $n$ times (same value for all observations in the leaf):
$$
\mathbf{x}_{\text{long}} = \begin{bmatrix} \:\mathbf{x}^k \: \mathbf{x}^k \:\: .. \:\: \mathbf{x}^k \: \end{bmatrix}
$$

With $\mathbf{x}^k$ being the underlying variables to solve for, which will be set as leaf values in gradient boosted trees.

Assuming that the Hessians are positive-definite (note that multinomial logistic Hessians are singular matrices), this can again be reformulated as a linear regression problem, this time of $k$ variables, given by a matrix $\mathbf{X}$ of dimensions $n k \times k$.

Denote the (upper) Cholesky or other square-rooting of the Hessians for each observation as:
$$
\mathbf{C}^{k \times k}_i = \mathbf{H}^\frac{1}{2}_i
$$

Then denote the vector of length $n$ made up by concatenating only the elements $(p,q)$ of the matrices $\mathbf{C}_i$ for every observation:
$$
\mathbf{c}_{p,q} = \begin{bmatrix}

c^1_{p,q} \: c^2_{p,q} \:\: .. \:\: c^n_{p,q}

\end{bmatrix}
$$

And the vector of lenght $n$ made up by concatenating only the elements $j$ of the vectors $\mathbf{v}_i = \mathbf{C}^{-1}_i \mathbf{g}_i$:
$$
\mathbf{z}_j = \begin{bmatrix}

v^1_j \: v^2_j \:\: .. \:\: v^n_j

\end{bmatrix}
$$

The solution of the gradient boosting problem is then equivalent to the linear regression problem of these variables:
\begin{equation}
\mathbf{X} = \begin{bmatrix}

\mathbf{c}_{1,1} \: \mathbf{c}_{1,2} \:\: .. \:\: \mathbf{c}_{1,k} \\

\mathbf{c}_{2,1} \: \mathbf{c}_{2,2} \:\: .. \:\: \mathbf{c}_{2,k} \\

.. \\

\mathbf{c}_{k,1} \: \mathbf{c}_{k,2} \:\: .. \:\: \mathbf{c}_{k,k}

\end{bmatrix} \:\:\:\:
\mathbf{y} = \begin{bmatrix}

\mathbf{z}_1 \\

\mathbf{z}_2 \\

.. \\

\mathbf{z}_k
\end{bmatrix}
\end{equation}

(In the case of Cholesky factorization, the vectors $\mathbf{c}_{i,j}$ with $i > j$ would be just zeros)

For practical purposes, it is easier to imagine operating on row-major arrays of dimensions $(n, k)$ for the gradients and $(n, k, k)$ for the Hessians, given the block structure of the problem.

The solution to the gradient boosting leaf problem can then be obtained in a simpler structure as follows:
$$
\mathbf{x}^{*} = \beta^{*} = \mathbf{A}^{-1} \mathbf{b}
$$

With
$$
\mathbf{A} = \sum_{i=1}^n \mathbf{H}_i \:\:\:\:\:\:\:\:
\mathbf{b} = \sum_{i=1}^n \mathbf{g}_{i}
$$

Denote the function values of the quadratic problem as:
$$
q(\mathbf{x}) = 0.5 \mathbf{x}^T \mathbf{A} \mathbf{x} - \mathbf{x}^T \mathbf{b}
$$

And note that $\mathbf{x}^{*} = \mathbf{A}^{-1} \mathbf{b}$ is the minimizer of $q(\mathbf{x})$.

The function values from this quadratic problem formulation are connected to the squared residuals of the equivalent linear regression formulation as follows:
$$
rss =
|| \mathbf{y} - \mathbf{X} \beta ||^2
= 2 q(\beta) + \sum_{i=1}^n \mathbf{g}_{i}^T \mathbf{H}^{-1}_i \mathbf{g}_{i}
$$

The terms $\mathbf{g}_{i}^T \mathbf{H}^{-1}_i \mathbf{g}_{i}$ do not depend on the variables being solved for, so they can be left out of the minization problem, but they are useful to understand the squared residuals used to determine split thresholds.

\section{The algorithm}
The algorithm described earlier for finding the optimal split threshold and leaf values can be thought of as a procedure that starts with all the observations assigned to the right branch, and performs a pass moving one observation (in sorted order) from the current right branch to the left branch.

Recall again that the aims of this procedure are to find the optimal split (assignment of observations to either the left branch or the right branch) as measured by minimizing squared residuals, and to determine the optimal values to set as prediction for each branch.

At a given split point, these will be given by:
$$
rss = rss_{\text{left}} + rss_{\text{right}} = 2 q_{\text{left}}(\mathbf{x}_{\text{left}}) + (\sum_{i \in L} \mathbf{g}_{i}^T \mathbf{H}^{-1}_i \mathbf{g}_{i}) + 2 q_{\text{right}}(\mathbf{x}_{\text{right}}) +  (\sum_{i \in R} \mathbf{g}_{i}^T \mathbf{H}^{-1}_i \mathbf{g}_{i})
$$

With:
$$
q_{\text{left}}(\mathbf{x}_{\text{left}}) = 0.5 \mathbf{x}_{\text{left}}^T (\sum_{i \in L} \mathbf{H}_i) \mathbf{x}_{\text{left}} - \mathbf{x}_{\text{left}}^T (\sum_{i \in L} \mathbf{g}_{i})
$$

Note that:
$$
(\sum_{i \in L} \mathbf{g}_{i}^T \mathbf{H}^{-1}_i \mathbf{g}_{i}) + (\sum_{i \in R} \mathbf{g}_{i}^T \mathbf{H}^{-1}_i \mathbf{g}_{i})
=
\sum_{i=1}^n \mathbf{g}_{i}^T \mathbf{H}^{-1}_i \mathbf{g}_{i}
$$

Hence, if one observation is taken away from the right branch and assigned to the left branch, this term will decrease on the right branch by the same value that it increases in the left branch, which for splitting purposes means it can be treated as a constant that does not depend on either the split threshold or the leaf values. Thus, the iterative sorted-order algorithm only needs to keep track of where the following quantity reaches its minimum:
$$
\min q_{\text{left}}(\mathbf{x}_{\text{left}}) + q_{\text{right}}(\mathbf{x}_{\text{right}})
$$

Also note that the $\mathbf{A}$ and $\mathbf{b}$ variables for each branch can be updated cumulatively, since:
$$
\mathbf{A}_{\text{right}} = \mathbf{A} - \mathbf{A}_{\text{left}} \:\:\:\:\:\:
\mathbf{b}_{\text{right}} = \mathbf{b} - \mathbf{b}_{\text{left}}
$$

With this, the optimal split point for a feature and the optimal leaf values can be obtained in an iterative procedure as follows:

\begin{algorithm}[H]
\caption{Vector sorted-indices split procedure}
\hspace*{\algorithmicindent}
    \textbf{Inputs}
        Gradients $\mathbf{g}_{i..n}$ and
        Hessians $\mathbf{H}_{i..n}$ in sorted order of feature $f$
\begin{algorithmic}[1]
\State Calculate initial aggregations:
$$
\mathbf{A} = \sum_{i=1}^{n} \mathbf{H}_i \:\:\:\:\:\: \mathbf{b} = \sum_{i=1}^{n} \mathbf{b}_i
$$
\State Initialize:
$$
\mathbf{A}_{\text{left}} = \mathbf{0}^{k \times k}  \:\:\:\:\:\: \mathbf{b}_{\text{left}} = \mathbf{0}^{k} \:\:\:\:\:\: qp_{\text{best}} = \infty
$$
\For {$1..(n-1)$}
    \State Update $\mathbf{A}_{\text{left}} \gets \mathbf{A}_{\text{left}} + \mathbf{H}_i$, $\mathbf{b}_{\text{left}} \gets \mathbf{b}_{\text{left}} + \mathbf{g}_i$
    \State Obtain $\mathbf{x}_{\text{left}} = \mathbf{A}_{\text{left}}^{-1} \mathbf{b}_{\text{left}}$ and $\mathbf{x}_{\text{right}} = (\mathbf{A} - \mathbf{A}_{\text{left}})^{-1} (\mathbf{b} - \mathbf{b}_{\text{left}})$
    \State Calculate
    $$
    qp = 0.5 \mathbf{x}_{\text{left}}^T \mathbf{A}_{\text{left}} \mathbf{x}_{\text{left}} - \mathbf{x}_{\text{left}}^T \mathbf{b}_{\text{left}} + 0.5 \mathbf{x}_{\text{right}}^T (\mathbf{A} - \mathbf{A}_{\text{left}}) \mathbf{x}_{\text{right}} - \mathbf{x}_{\text{right}}^T (\mathbf{b} - \mathbf{b}_{\text{left}})
    $$
    \If {$qp < qp_{\text{best}}$}
        \State $qp_{\text{best}} \gets qp$ (update best residual)
        \State $t \gets i$ (update best threshold)
        \State $\mathbf{x}^{*}_{\text{left}} \gets \mathbf{x}_{\text{left}}$, $\mathbf{x}^{*}_{\text{right}} \gets \mathbf{x}_{\text{right}}$ (update leaf values)
    \EndIf
\EndFor
\Return $t$, $\mathbf{x}^{*}_{\text{left}}$, $\mathbf{x}^{*}_{\text{right}}$
\end{algorithmic}
\end{algorithm}

While the algorithm implies solving a linear system of dimension $k$ at each iteration, in practice, when using the histogram method, only a subset of possible split points are considered. Note that, if one were to skip some possible thresholds from consideration, the procedure would only need to sum values of $\mathbf{H}_i$ and $\mathbf{g}_i$ for the observations whose thresholds were skipped in order to resume at a subsequent threshold. Typically, histograms in commonly used GBT libraries consider at most 255 possible thresholds by default, and typical values of $k$ rarely exceed more than 15 or so for objectives of interest, with small values like $k=3$ being more common. Hence, for the case of histograms, assuming that the number of observations is comparatively larger, the linear solves and evaluations should be comparatively much faster than summing the required quantities on each histogram bin. Note that, since the $\mathbf{H}_i$ matrices are symmetric, it's only necessary to sum $\frac{k (k + 1)}{2}$ elements out of them from each observation instead of $k^2$.

Note also that for these calculations, one may reuse the Cholesky factorization of $\mathbf{A}$ for both the inverse product and the criterion calculation:
$$
\mathbf{x}^T \mathbf{A} \mathbf{x} = \mathbf{x}^T (\mathbf{L} \mathbf{L}^T) \mathbf{x} = || \mathbf{L}^T \mathbf{x} ||^2
$$

In practice, if the amount of data to aggregate is too large, for better computational accuracy on floating point numbers of limited precision, one might prefer to calculate means instead of sums, multiplying the values in the $qp$ criterion by the fraction of observations on each branch, and adjusting the computations of $\mathbf{A}_{\text{right}}$ and $\mathbf{b}_{\text{right}}$ accordingly by these fractions.

\section{Multinomial logistic case}
The algorithm sketched in the previous section is applicable to functions with positive-definite Hessians.

Consider again the case of multinomial logistic log-likelihood as objective. It produces probabilities of belonging to each of $k$ possible classes based on a vector of predictions as follows:
$$
P(y = c) = \frac{ \exp x_{c} }{ \sum_{i=1}^k \exp x_i }
$$

The gradient for the negative log-likelihood under these probabilities is given by:
$$
\mathbf{g} = I[y = c] - \mathbf{p}
$$

Where $\mathbf{p}$ is the vector of predicted probabilities, and $I[y = c]$ is an indicator function vector which has a value of 1 where the condition $y=c$ is met and a zero elsewhere. Note that $\sum_{i=1}^k p_i = 1$

The Hessian is given by:
$$
\mathbf{H} = \text{diag}(\mathbf{p}) - \mathbf{p} \mathbf{p}^T
$$

This Hessian will have a linear dependency, with one of its eigenvalues being zero - i.e. the matrix is singular. Hence, it cannot be inverted.

Note that:
$$
\frac{ \exp (x_{c} + k) }{ \sum_{i=1}^k \exp (x_i + k) }
=
\frac{ \exp x_{c} \exp k }{ \sum_{i=1}^k (\exp x_i  \exp k) }
=
\frac{ \exp x_{c} \exp k }{ \exp k \sum_{i=1}^k \exp x_i }
=
\frac{ \exp x_{c} }{ \sum_{i=1}^k \exp x_i }
$$

Where $k$ is a constant.

This means that the objective function is invariant to additions and substractions by the same constant, and different input vectors $\mathbf{x}$ can produce the same function values. For instance, if given an optimal solution $\mathbf{x}^{*}$, then any other vector $\mathbf{x}^{**} = \mathbf{x}^{*} + k$ will be equally as optimal, which goes along with its Hessian not being positive-definite.

From this, it can be seen that only $k-1$ values are estimable. Typically, in linear models with multinomial logistic log-likelihood, one of the classes is set as the reference class, which is equivalent to constraining its value $x_{\text{ref}}$ to always be zero. Thus, for practical purposes, one might drop the last category for instance, setting that prediction as always zero and not updating its value with boosting iterations. Then, this implies operating on variables $\mathbf{g}$ and $\mathbf{H}$ of dimensions $k-1$ and $(k-1) \times (k-1)$ only, for example by ignoring the last element of each vector $\mathbf{p}$ in the calculations:
$$
\mathbf{H}_{\text{reduced}} = \text{diag}(\mathbf{p}_{1..(k-1)}) - \mathbf{p}_{1..(k-1)} \mathbf{p}_{1..(k-1)}^T
$$

For this reduced problem with a reference category, one might likewise initialize GBT predictions with intercepts where the last element is constrained to be zero, by first calculating the logarithms of the proportions of each class:
$$
\mathbf{x}^0 = \log \begin{pmatrix} \frac{1}{n} \sum_{i=1}^n I[y_i = 1] \:\:\: \frac{1}{n} \sum_{i=1}^n I[y_i = 2] \:\:\:\: .. \:\:\:\: \frac{1}{n} \sum_{i=1}^n I[y_i = k] \end{pmatrix}
$$

And then subtracting the last value, so that its prediction will start off as zero:
$$
\mathbf{x}^0 \gets \mathbf{x}^0 - x^0_k
$$

However, when regularization is applied to the GBT predictions, the Hessian is no longer singular:
$$
\text{diag}(\mathbf{p}) - \mathbf{p} \mathbf{p}^T + \lambda I^{k \times k}
$$

And shifting a vector $\mathbf{x}^{*}$ by a constant would not change the probabilities, but would change the regularization values, which would need to be taken into account in the criterion to minimize.

Thus, under regularization, all $k$ components are estimable and must be taken into account in the aggregations.

In either of these cases, for more efficient computations, since the Hessians consist of a sum of diagonals and a symmetric rank-n update, one might keep track of these by aggregating separately:
$$
\mathbf{d} \gets \mathbf{d} + \mathbf{p}_i \:\:\:\:\:\: \mathbf{R} \gets \mathbf{R} + \mathbf{p} \mathbf{p}^T \:\:\:\:\:\: \mathbf{v} \gets \mathbf{v} + I[y_i = 1]
$$

(with only the upper or lower triangular part of $\mathbf{R}$ being updated)

Then:
$$
\mathbf{A} = \text{diag}(\mathbf{d}) - \mathbf{R} \:\:\:\:\:\: \mathbf{b} = \mathbf{v} - \mathbf{d}
$$

\section{Comparison against alternatives}
The XGBoost library introduced a procedure for gradient boosting on vector functions in version 2.0\footnote{\url{https://github.com/dmlc/xgboost/blob/307b87333cd3c86cbfb76c60ea392797c74e9bef/NEWS.md}}, which is based on a diagonal upper bound of the Hessians, very similarly to \cite{glmnet}.

In their procedure, a diagonal upper bound which diagonally dominates the true Hessian is used:
$$
\mathbf{ \tilde{H} } = 2 \: \text{diag}(\mathbf{p})
$$

As this matrix is diagonal and there are no dependencies, the leaf estimation problem is equivalent to a multi-task linear regression problem as in \cite{glmnet}, with the difference that in XGBoost's case, each task has a different weight given by the corresponding elements of this upper-bounded Hessian.

This makes estimation easy as leaf values are just weighted means of gradients divided elementwise by these upper-bounded Hessians, and the gain criterion is likewise a sum of sums, but using this matrix in place of the true Hessian in a second-order approximation results in a worse approximation to the original function.

As a practical experiment, the two approaches were compared in what's perhaps the simplest possible task: estimating optimal intercepts by gradient boosting iterations for multinomial logistic log-likelihood, starting from zero, with the same prediction for all observations. For this purpose, the first 5000 observations of the 'Cover Type' dataset\footnote{\url{https://archive.ics.uci.edu/dataset/31/covertype}} were used. This problem can be reduced by grouping all observations with the same class into a single observation weighted by the class frequencies (since no features are used and there are no previous raw scores), and it allows an easy closed-form solution, but can nevertheless serve as a basis for comparison.

In this simple experiment, it took 4 iterations for the method with true Hessians to converge to within $10^{-6}$ of the optimal value (as measured by multinomial log-likelihood), while it took over 400 iterations for the diagonal upper bound approach to converge to the same level:

\includegraphics[totalheight=6cm]{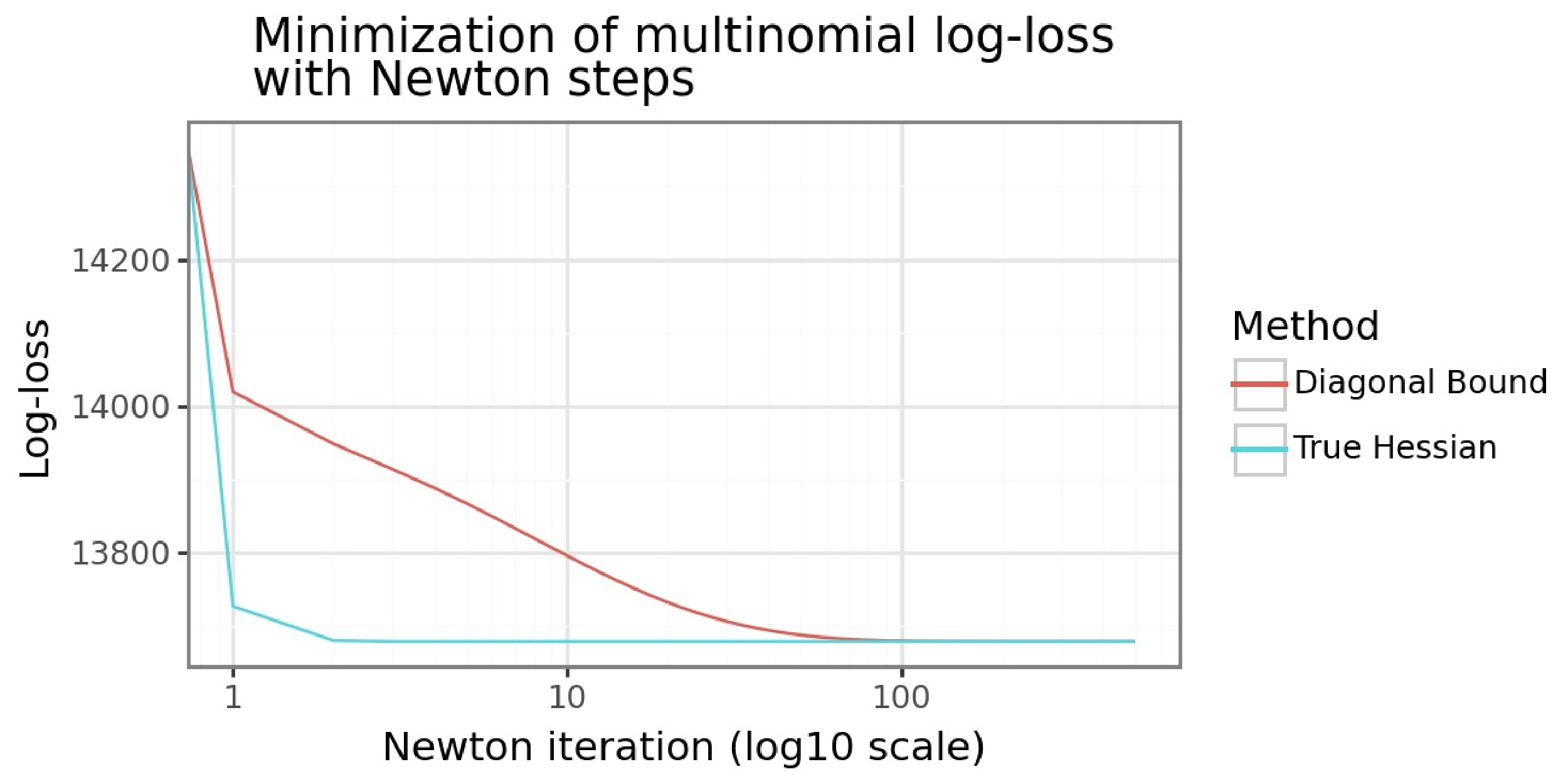}

While perhaps attractive in theory, the fact that it takes so many iterations for the diagonal approach to converge in such a simple task is perhaps worrysome for practical applications, where the number of boosting iterations in total might be less than that.

The source code for the experiment and for a proof-of-concept of the methodology is made open-source and publicly available\footnote{\url{https://github.com/david-cortes/vector_gbt_poc/blob/master/Example1_Calculating_Intercepts.ipynb}}.

As a second experiment, this was followed up by finding a split threshold and leaf values based on the first column in the data. In the case of XGBoost, it was done by manually setting the parameters to reflect just the basic framework here - no regularization, unit step size, a single split (depth=1), only the first column in the data being considered, number of histogram bins large enough that every possible threshold would be considered, and no sub-sampling of any kind. The split point selected by XGBoost in this case matched exactly with the split point that minimizes the RSS criterion:

\includegraphics[totalheight=6cm]{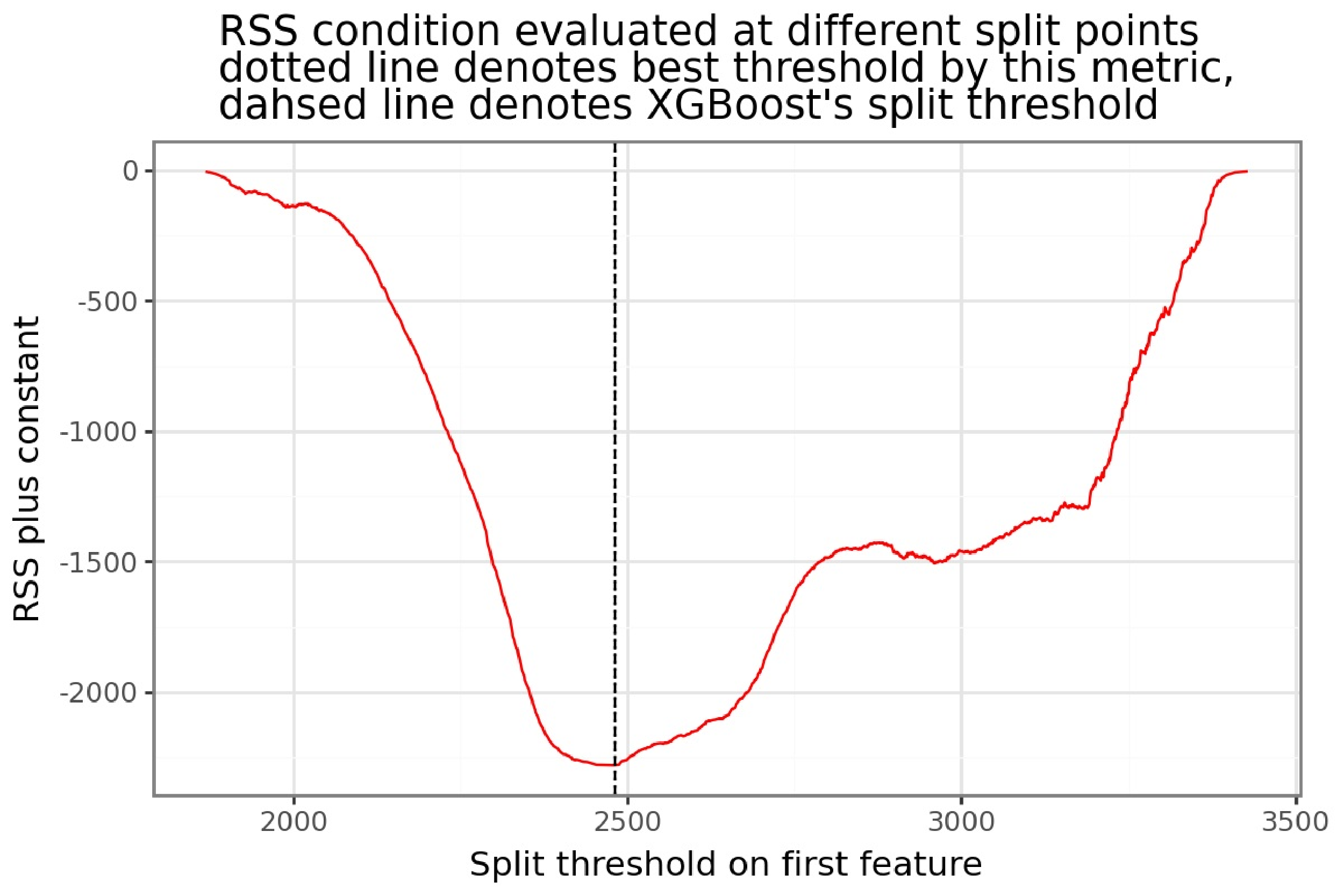}

However, the leaf values differ, which makes the two still end up with different values of the objective function being minimized, with the leafs calculated from the true Hessian faring better in this regard:

\includegraphics[totalheight=6cm]{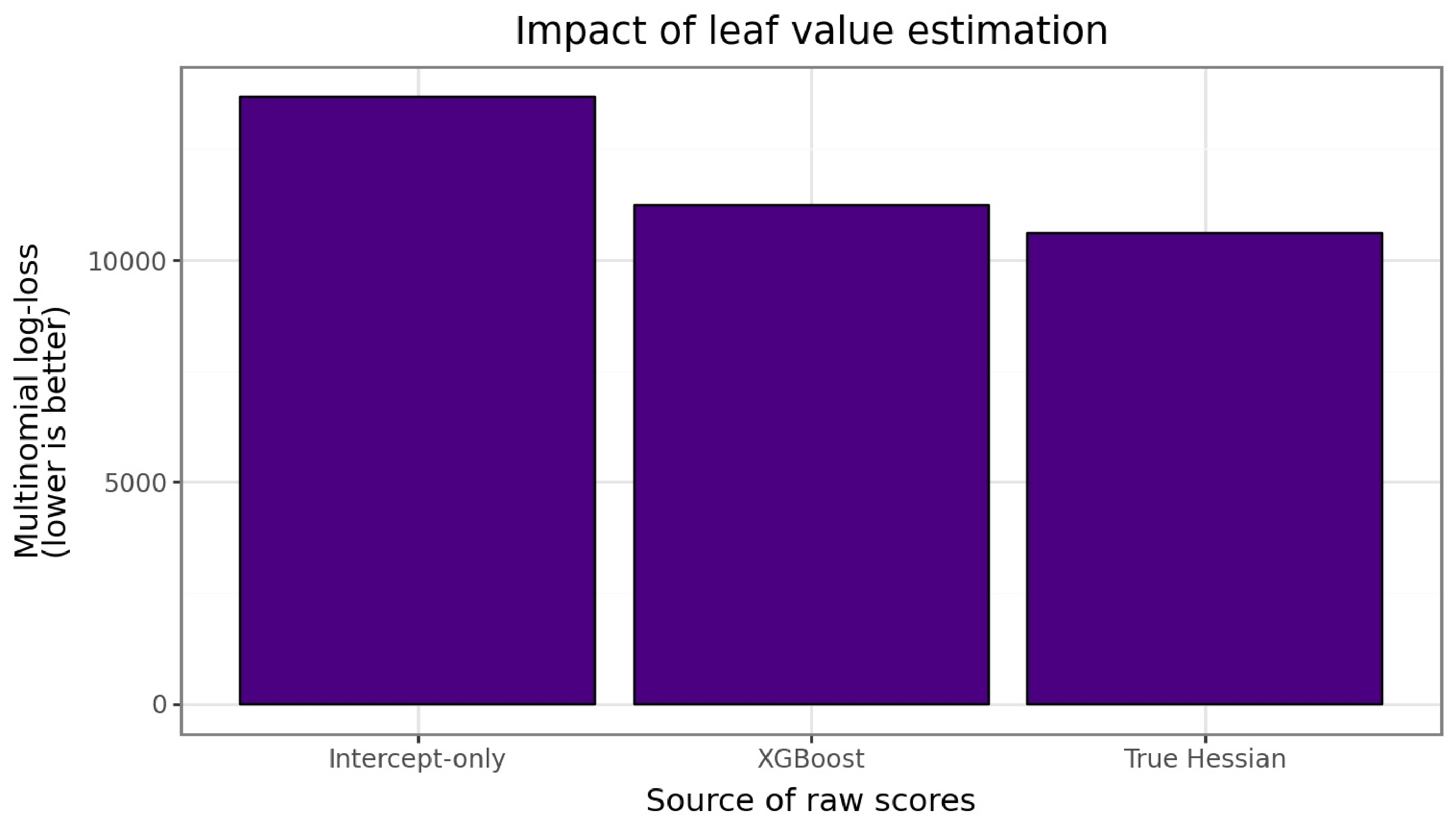}

Whether the splits would show large differences in other scenarios still remains to be tested.

\section{Conclusions and future work}

This work introduced a logical extension of the gradient boosting framework to functions of vectors using their full Hessians, based on simple calculus and linear algebra and aided by vector GLMs for explanations and derivations. This usage of true second derivatives should result in better approximations at the cost of quadratic scalability with the dimensionality of the vectors. Its advantages over diagonal approximations were showcased in a very simple example with real data, but it still remains to implement a full gradient boosting procedure based on these ideas and compare full GBT results in real data.

\bibliographystyle{plain}
\bibliography{gbt_multi}

\end{document}